# Use of Artificial Intelligence to Analyse Risk in Legal Documents for a Better Decision Support

Dipankar Chakrabarti
*Senior Member, IEEE*
*PricewaterhouseCoopers Pvt Ltd.*
Kolkata, India
dipankar.chakrabarti@pwc.com

Neelam Patodia
*PricewaterhouseCoopers Pvt Ltd.*
Kolkata, India
neelam.patodia@pwc.com

Udayan Bhattacharya
*PricewaterhouseCoopers Pvt Ltd.*
Kolkata, India
udayan.bhattacharya@pwc.com

Indranil Mitra
*PricewaterhouseCoopers Pvt Ltd.*
Kolkata, India
mitra.indranil@pwc.com

Satyaki Roy
*PricewaterhouseCoopers Pvt Ltd.*
Kolkata, India
satyaki.roy@pwc.com

Jayanta Mandi
*PricewaterhouseCoopers Pvt Ltd.*
Kolkata, India
jayanta.mandi@pwc.com

Nandini Roy
*PricewaterhouseCoopers Pvt Ltd.*
Kolkata, India
nandini.roy@pwc.com

Prasun Nandy
*PricewaterhouseCoopers Pvt Ltd.*
Kolkata, India
prasun.nandy@pwc.com

*Abstract*—Assessing risk for voluminous legal documents such as request for proposal, contracts is tedious and error prone. We have developed "risk-o-meter", a framework, based on machine learning and natural language processing to review and assess risks of any legal document. Our framework uses Paragraph Vector, an unsupervised model to generate vector representation of text. This enables the framework to learn contextual relations of legal terms and generate sensible context aware embedding. The framework then feeds the vector space into a supervised classification algorithm to predict whether a paragraph belongs to a pre-defined risk category or not. The framework thus extracts risk prone paragraphs. This technique efficiently overcomes the limitations of keyword based search. We have achieved an accuracy of 91% for the risk category having the largest training dataset. This framework will help organizations optimize effort to identify risk from large document base with minimal human intervention and thus will help to have risk mitigated sustainable growth. Its machine learning capability makes it scalable to uncover relevant information from any type of document apart from legal documents, provided the library is pre-populated and rich.

## I. INTRODUCTION

A contract between two parties defines the scope of work and commercial business terms for performing such activities. It is very important for any business organization to review the contract and analyze risks, such as liability, indemnity, risk purchase and other such commercial risks. Early identification of risks help either to mitigate the risks or to take a decision of entering into contract understanding risk-reward ratio. Organizations traditionally rely upon manual reading by legal professionals to assess risks emanating out of the documents. The continued influx of legal paperwork demands more of the lawyer's time and knowledge/experience for review. This time consuming, cost intensive and person dependent activity is riddled with inefficiencies. Even after investing 11.2 hours per week [1] in document creation and management, chances of error still persist because of unidentified or misinterpreted risk aspects, which could interfere with an organization's performance while increasing financial risk.

Thus there is an increased demand for intelligently automating analysis of contracts and other legal documents and to provide correct interpretation with minimum intervention of human beings. This is far beyond a "contract management system" which files and indexes electronic contracts/legal documents. The focus of this paper is to propose a contract analysis system efficient at identifying and highlighting embedded risks in contracts or other legal documents.

Traditional keyword driven approach for contract risk analysis does not capture the contextual understanding of different clauses which limits its performance in the following two ways: (1) identifying paragraphs which contain any of the library keywords as risk prone, thus raising false alarms, (2) understanding the risk significance of a keyword in context of another keyword.

We have developed an effective and intelligent framework named "risk-o-meter" based on machine learning (ML) and natural language processing (NLP). Our framework dramatically changes the way contractual risks are assessed by identifying risk prone paragraphs and associating them to their predefined risk categories like liability, indemnity, confidentiality and other such commercial risks. It reduces manual effort and operational time; increases consistency of outcomes; enhances precision in risk identification and reduces chances of overlooking critical information through manual fatigue or inexperience of reviewer. It would thus help in creating a risk-aware environment for sustainable growth of organizations.

The paper is organized as follows: work done in the field of word embeddings is given in Section II. We then present the building blocks of our "risk-o-meter" in Section III, and particularly focus on the algorithms used for risk identification in Section IV. Following that we present an experimental comparison to assess most effective model for our task in Section V. We provide future possibilities in Section VI and conclusion in Section VII.

## II. RELATED WORK

Vector space models have been used in distributional semantics for quite some time. The term word embeddings was coined by Bengio et al. [17]. Colbert and Weston et al. [16] showed the utility of word embeddings and their usefulness for downstream tasks such as parsing, tagging, named entity recognition etc. Socher et.al [18] focused on distributed representation of phrases and sentences by implementing parsing techniques. Their method was supervised and required labelled data.

The unsupervised word vector model proposed by Mikolov et.al, 2013 [20] is an efficient method for learning high-quality distributed vector representations that capture a large number of precise syntactic and semantic word relationships. This model was extended beyond word-level to achieve paragraph-level representation. The framework proposed by Quoc Le and Tomas Mikolov, 2014 [2] for text representation applies () in learning vector representations from variable length pieces of text.

## III. BUILDING BLOCKS OF RISK-O-METER

The objective of our framework "risk-o-meter" is to identify risk prone paragraphs from legal documents that belong to a predefined risk category. Risk categories are defined as liability, confidentiality, indemnity, termination and others. The categories can be modified/updated/added as needed. The framework consists of four key blocks: (1) Training Data, (2) Text Representation, (3) Text Classification, (4) Continuous Learning (see Figure. 1).

### A. Training Data

The training dataset is created by tagging such paragraphs collated from contractual/legal documents to their respective risk categories. The categories are generated by human annotators. In case a paragraph belonged to more than one category, it is recorded separately for all these categories. In our training dataset, we consider all the risk prone paragraphs and their associated categories uniformly. The quality and quantity of the training data collated, has a significant impact on the performance of our framework. The specifics of this impact is described in detail in Section V.

### B. Text Representation

Text representation is used to convert text into a machine readable format. In order to represent risk prone paragraphs, it is essential to understand and capture the context in which the legal terms are used i.e. their meaning. E.g. "The agency shall indemnify the department against all third-party claims of infringement of copyright, patent, trademark or industrial design rights arising from use of the Goods or any part thereof in India". In this example, indemnify used in the context of agency, department, third party and infringement essentially qualifies as risk.

The commonly used bag-of-words (BoW) model [2] and its extension, term frequency-inverse document frequency model (TF-IDF) [15], though quite simple and efficient cannot be used in our case. It suffers from two main disadvantages: (1) it loses the ordering of the words, as a result sentences having the same words are represented in an identical manner, (2) it also ignores the semantics of the words, meaning it does not take into consideration the distance between words.

In our framework, we use Paragraph vectors proposed in [2] for learning high quality, continuous distributed vector representations that capture a large number of precise syntactic and semantic relationship between words and the topic of the paragraph. This n-dimensional vector space is created for each paragraph in the training dataset. After the training converges, these feature vectors are used for calculating paragraph vectors for the unseen documents. The specifics of this model is described in detail in Section IV-A.

### C. Text Classification

Vector representations from text representation module are then fed to the text classification module to predict whether a paragraph belongs to a particular risk category or not. While cosine similarity measures are helpful in predicting the categories; in this paper, we implement supervised learning techniques, namely: Support Vector Machines and Naïve Bayes, to further train the system to predict better. The probability values generated post classification for unseen documents depict the likelihood of the paragraph being associated with the concerned risk category. Both these models are effective in handling high dimensional vector spaces and have been detailed in Section IV-B.

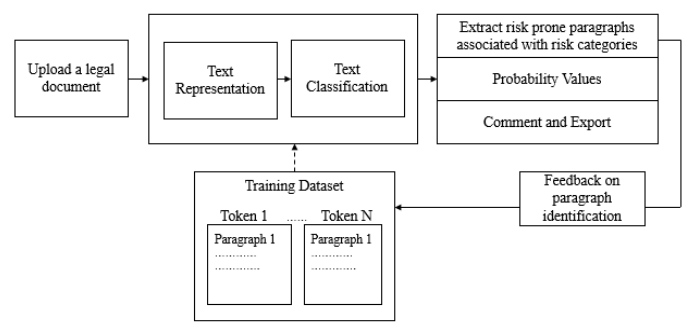

Fig.1. Process flow diagram for our AI enabled framework, "risk-o-meter"

### D. Continuous Learning

Our framework has an integrated feedback loop which records review responses in the form of acceptance or rejection for all the identified paragraphs for a given risk category and appends them to the training data. Unidentified clauses can be manually added to the training data. This updated training data is later used to retrain the models. This way the machine learns continuously and performs better.

## IV. SPECIFICATIONS OF RISK IDENTIFICATION

In this section we will detail the algorithm behind Paragraph Vectors and classification techniques: Support Vector Machines and Naïve Bayes, as well as define their corresponding performance metrics.

### A. Paragraph Vectors

Paragraph vector, an unsupervised neural network model generates sensible context aware word embedding for input sentences of variable length [2]. It has only one hidden layer. It does not rely upon parse trees. It is an extension of the word vector model [20].

In the word vector model, only words are considered as input nodes. However, in Paragraph vector, each paragraph id also acts as an input node and is mapped to a unique vector. We have assigned a unique label or paragraph id for each paragraph as the meaning of the paragraphs vary even if they belong to the same category. The paragraph vector (D) along with word vector (W) is considered as a member of the context set. The context is sampled from a sliding window of fixed length over the paragraph. The context window considers words to the left and right of the target word. While the paragraph vector is shared only across the context from the same paragraph, the word vectors are universally shared across all paragraphs [2]. The concatenation or sum of the vectors (W and D) is then used to predict the next word in the context.

*1) Hyperparameter selection:* Building this network for a given learning task involves selecting optimal hyper-parameters. Hyper-parameter choice is also crucial for an improved performance (both accuracy and speed). These are as follows:

 a) Choosing between the two neural networks based models: Distributed memory (PV-DM) and Distributed bag of words (PV-DBOW). Given a window of words $\{w_1, w_2, w_3, w_4, w_5\}$, the DM model predicts $w_3$ given the rest, while the DBOW model predicts $w_1, w_2, w_4, w_5$ given $w_3$.

 b) Choosing the algorithm for training the selected model: hierarchical softmax (HS) and negative sampling (NEG).

 c) Dimensionality of the feature vectors.

 d) Window size which determines the maximum distance between the current and predicted word within a paragraph.

 e) Minimum frequency so that all words lower than this threshold is ignored.

 f) Concatenation vs sum/average of context vectors.

 g) Sample threshold so that high frequency words are randomly down-sampled.

*2) Training Procedure:* The first task in building paragraph vectors is determining the efficacy of the base architecture against the training algorithms for the given dataset. We evaluated the performance of both the models: PV-DM and PV-DBOW as shown in Table 1. DM trained using negative sampling outperforms the other combinations DBOW-HS, DBOW-NEG and DM-HS at the task of identifying the risk prone paragraphs. We thus opted for the DM model using negative sampling to generate vector representations.

TABLE I. END-TO-END PERFORMANCE COMPARISON AMONGST TRAINING ARCHITECTURE AND ALGORITHM TO PREDICT THE RISK CATEGORY (TERMINATION-HAVING MAXIMUM EXAMPLES IN THE TRAINING DATA). INITIAL MODEL PARAMETERS: NEGATIVE SAMPLE (K) OF 5, SUBSAMPLING (T) OF $10^{-6}$, CONTEXT WINDOW=5, VECTOR SIZE=300, SUPPORT VECTOR MACHINES (SVM) LINEAR CLASSIFIER WITH C VALUE OF 1

| Risk Category | Method | Accuracy | Precision | Recall | F1-score |
|---|---|---|---|---|---|
| Termination | **DM-NEG** | **87%** | **74%** | **84%** | **79%** |
| | DM-HS | 81% | 86% | 43% | 58% |
| | DBOW-HS | 87% | 96% | 59% | 73% |
| | DBOW-NEG | 87% | 96% | 59% | 73% |

The objective of the PV-DM model is to maximize the average log probability, given by:

$$\frac{1}{T}\sum_{t=n}^{T-n} \log p(w_t | w_{t-n}, \dots, w_{t+n}) \quad (1)$$

where, $w_t$ represents the target word and $w_{t-n}$ to $w_{t+n}$ represents the input context words ($w_O$) with a window of n words at each time step t. T represents the sequence of words $w_1, w_2, \dots, w_T$ in the given training set that belong to a vocabulary V ( T $\subset$ V) whose size is |V|.

The softmax layer calculates this probability as:

$$p(w_I | w_O) = \frac{exp(v'_{w_I}{}^T v_{w_O})}{\sum_{w_i=1}^{V} exp(v'_{w_i}{}^T v_{w_O})} \quad (2)$$

where, $v_w$ and $v'_w$ are the input and output vector representations of word w; $w_I$ represents the i<sup>th</sup> target word.

Computing the softmax is expensive as the inner product between $v_w$ and the output embedding $v'_w$ of every word $w_i$ in the vocabulary V needs to be computed as part of the sum in the denominator in order to obtain the normalized probability of the target word given its context.

Negative sampling on the other hand, is similar to stochastic gradient descent: instead of changing all of the weights each time with taking into account all of the thousands of observations, we're using only sample (K) of them and increasing computational efficiency dramatically too [6]. The objective of negative sampling for one observation is as follows:

$$\log p(w_I | w_O) = \log \sigma(v'_{w_I}{}^T v_{w_O}) + \sum_{i=1}^{K} E_{w_i \sim P_n(w)} [\log \sigma(-v'_{w_i}{}^T v_{w_O})] \quad (3)$$

The noise distribution $P_n(w)$ is defined as the Unigram distribution raised to the power of ¾.

$$P(w_i) = \frac{f(w_i)^{\frac{3}{4}}}{\sum_{j=0}^{n} \left( f(w_i)^{\frac{3}{4}} \right)} \quad (4)$$

where, *3/4* is the empirical value suggested in [2]; *f(w)* is the frequency of the word in the corpus.

TABLE II. END-TO-END PERFORMANCE COMPARISON OF PREDICTING THE RISK CATEGORY (TERMINATION) FOR VARYING VALUES OF K. THE ADDITIONAL MODEL PARAMETERS: SUBSAMPLING OF $10^{-6}$, CONTEXT WINDOW=5, VECTOR SIZE=300, SVM LINEAR CLASSIFIER WITH C VALUE OF 1

| Risk Category and Method | No. of samples K | Accuracy | Precision | Recall | F1-score |
|---|---|---|---|---|---|
| Termination with DM-NEG | 5 | 87% | 74% | 84% | 79% |
| | **10** | **88%** | **76%** | **86%** | **81%** |
| | 15 | 86% | 72% | 86% | 78% |
| | 20 | 85% | 70% | 86% | 78% |

Our experimental results (Table II) indicate that while Negative Sampling achieves a respectable accuracy even with K = 5, using K = 10 achieves considerably better performance with an accuracy of 88% and F1-score of 81%.

TABLE III. END-TO-END PERFORMANCE COMPARISON OF PREDICTING THE RISK CATEGORY (TERMINATION) FOR VARYING SUBSAMPLING THRESHOLDS. THE ADDITIONAL MODEL PARAMETERS: CONTEXT WINDOW=5, VECTOR SIZE=300, SVM LINEAR CLASSIFIER WITH C VALUE OF 1

| Risk Category and Method | Subsampling Threshold (T) | Accuracy | Precision | Recall | F1-score |
|---|---|---|---|---|---|
| Termination with DM-NEG (k=10) | 0 | 71% | - | 0% | - |
| | $10^{-5}$ | 89% | 86% | 73% | 79% |
| | **$10^{-6}$** | **88%** | **76%** | **86%** | **81%** |

Subsampling has also been used to counter the balance between rare and frequent words. Words like "is", "an", "the" and such similar stop words occur innumerable times in the dataset and do not provide valuable information as compared to the rare words. The vector representations of frequent words remain almost constant after training on several examples. E.g. Co-occurrences of "Limitation" and "Liability" hold much more significance than "The" and "Liability", as mostly all words in a paragraph co-occur with such words. Thus words whose frequency, $f(w_i)$ is greater than the threshold, $T$, is subsampled using the given equation:

$$P(w_i) = 1 - \sqrt{\frac{T}{f(w_i)}} \quad (5)$$

As depicted in Table III, subsampling significantly improves the performance of our framework and a threshold value of $10^{-6}$ is optimal for our case. It also improves the training speed by nearly 8 times (for our training dataset).

The context window and vector size also play a significant role in the model's performance. Context window of 10 (Table IV) and vector size of 100 (Table V) are optimal for our case. We ignored all words from the corpus which had a frequency lesser than 5. In our paper, we have used concatenation to combine the two vectors as it keeps the ordering information intact.

TABLE IV. END-TO-END PERFORMANCE COMPARISON OF PREDICTING THE RISK CATEGORY (TERMINATION) FOR VARYING CONTEXT WINDOW SIZE. THE ADDITIONAL MODEL PARAMETERS: K=10, T=$10^{-6}$, VECTOR SIZE=300, SVM LINEAR CLASSIFIER C-VALUE=1

| Risk Category and Method | Window Size | Accuracy | Precision | Recall | F1-score |
|---|---|---|---|---|---|
| Termination with DM-NEG | 5 | 88% | 76% | 86% | 81% |
| | 8 | 88% | 77% | 84% | 80% |
| | **10** | **90%** | **82%** | **84%** | **83%** |

TABLE V. END-TO-END PERFORMANCE COMPARISON OF PREDICTING THE RISK CATEGORY (TERMINATION) FOR VARYING VECTOR SIZE. THE ADDITIONAL MODEL PARAMETERS: K=10, T=$10^{-6}$, CONTEXT SIZE=10, SVM LINEAR CLASSIFIER C-VALUE=1

| Risk Category and Method | Vector Size | Accuracy | Precision | Recall | F1-score |
|---|---|---|---|---|---|
| Termination with DM-NEG | **100** | **92%** | **90%** | **82%** | **86%** |
| | 200 | 91% | 89% | 77% | 83% |
| | 300 | 90% | 82% | 84% | 83% |

This neural network based vector model is trained using stochastic gradient descent where the gradient is obtained via backpropagation.

TABLE VI. 5 MOST SIMILAR WORDS ASSOCIATED WITH THE TARGET WORD

| Target Word | Lower Case | Without Case Conversions |
|---|---|---|
| Termination | notice | contract |
| | contract | prejudice |
| | order | written |
| | date | notice |
| | than | whole |
| Indemnity | infringements | Trademark |
| | damages | alleged |
| | alleged | attorney's |
| | losses | nature |
| | claims | suits |
| Insurance | agrees | coverage |
| | secure | taken |
| | contribution | commencing |
| | employee's | policies |
| | place | place |

Once the optimal hyper-parameters are selected, we also qualitatively evaluate these word embeddings by inspecting manually the five most similar words (by cosine similarity) [19] to a given set of target words. We compare the results across two cases: converting all words to lower cases and without making any case conversions. It is evident from Table VI that PV-DM model finds words that associate with the target word (domain aspect). It also depicts that paragraph vector model draws out almost similar logical associations for the both the cases. We also compared their end to end results and found that they were similar to Table V for vector size 100. We prefer to adopt the lower case model as it is more robust.

*3) Inference:* After the training converges, these feature vectors are used for calculating paragraph vectors for the unseen documents, where-in all weights are fixed. We retrain the model with words present in the unseen documents, but it does not impact the members in the context set.

## B. Classification techniques

We opted for implementing both Support Vector Machines (SVM) and Naïve Bayes (NB) to classify the vector space based on their risk categories. As discussed earlier, these two techniques are chosen as they are efficient in handling high dimensions. We built individual classifiers for each of the predefined risk categories and compared their performance across both the classifier algorithms. The feature vectors are normalized before implementing classification algorithms to enable better projection. We built four types of classifier models: (1) SVM with Linear kernel, (2) SVM with Radial Basis Function (RBF) kernel, (3) Gaussian NB, (4) Bernoulli NB. Our results show that Support Vector Machines with linear kernel outperformed all other classifier models across all parameters (Table VII). The details of both the algorithms are as follows (for further details, refer [9, 13]):

TABLE VII. COMPARING THE PERFORMANCE OF VARIOUS CLASSIFIERS TO PREDICT THE RISK CATEGORY (TERMINATION) WHOSE VECTOR SPACE IS BUILT USING DM-NEG WITH K=10, T=$10^{-6}$, CONTEXT WINDOW=10 AND VECTOR SIZE=100

| Classifier Method | AUC | Accuracy | Precision | Recall | F1-score |
|---|---|---|---|---|---|
| **SVM-Linear** | **0.96** | **92%** | **90%** | **82%** | **86%** |
| SVM-Radial Basis Function | 0.96 | 71% | - | 0% | - |
| NB-Gaussian | 0.75 | 75% | 65% | 34% | 45% |
| NB-Bernoulli | 0.94 | 89% | 83% | 80% | 81% |

*1) Support Vector Machines:* SVM are based on the concept of decision planes that define decision boundaries. In the case of n dimensional space of input variables, a hyperplane splits positive and negative sets of examples, with maximal margin. C-value, a regularization parameter affects the number of instances that fall within the margin and influences the number of support vectors used by the model [9]. A large value of C permits more violations of the hyper plane and results in lesser sensitivity and higher bias. We cross-validated our model at different C-values as depicted in Table VIII.

TABLE VIII. COMPARING THE PERFORMANCE OF SVM-LINEAR CLASSIFIER TO PREDICT THE RISK CATEGORIES BY ADJUSTING DIFFERENT C VALUES

| Risk Category | C value | AUC | Accuracy | Precision | Recall | F1-score |
|---|---|---|---|---|---|---|
| Termination | 0.1 | 0.96 | 76% | 100% | 18% | 31% |
|  | **1** | **0.96** | **92%** | **90%** | **82%** | **86%** |
|  | 10 | 0.96 | 90% | 82% | 84% | 83% |
|  | 100 | 0.96 | 89% | 78% | 86% | 82% |
| Indemnity | 0.1 | 0.93 | 88% | - | 0% | - |
|  | **1** | **0.93** | **94%** | **91%** | **56%** | **69%** |
|  | 10 | 0.95 | 91% | 61% | 78% | 68% |
|  | 100 | 0.96 | 90% | 56% | 83% | 67% |
| Insurance** | 0.1 | 0.91 | 90% | - | 0% | - |
|  | 1 | 0.89 | 90% | - | 0% | - |
|  | 10 | 0.9 | 90% | - | 0% | - |
|  | 100 | 0.91 | 90% | - | 0% | - |

** The classifier for the risk category "Insurance", could not be optimized due to insufficient data.

*2) Naïve Bayes :* NB Classifier is a linear classifier built on the Bayesian theorem. It is based on the assumption that features in a dataset are mutually independent and are identically distributed. An additional assumption is the conditional independence of features.

The classifier's performance is evaluated in terms of area under the curve (AUC), accuracy, precision, recall and F1-score. A random classifier has an AUC of 0.5 while a perfect classifier has an AUC of 1.

TABLE IX. TEST RESULTS FOR THE FINAL MODEL PARAMETERS: DM-NEG, K=10, T=$10^{-6}$, CONTEXT WINDOW=10, VECTOR SIZE=100, SVM LINEAR CLASSIFIERS WITH OPTIMAL C VALUES

| Risk Category | Accuracy | Precision | Recall | F-score |
|---|---|---|---|---|
| Termination | 91% | 89% | 77% | 83% |
| Indemnity | 93% | 83% | 56% | 67% |

## V. RESULTS

The dataset on which this experimentation was performed consisted of three sets: 1,382 paragraphs for training, 151 paragraphs for validation and 75 paragraphs for test. These paragraphs were collated from request for proposal (RFP) documents of various risk categories. The risk categories with maximum number of paragraphs in the training dataset are: "Termination" with 499, "Indemnity" with 117 and "Insurance" with 89. In the validation dataset, Termination accounted for 44 paragraphs, Indemnity accounted for 18 paragraphs and Insurance accounted for 15 paragraphs. In the test dataset the count of paragraphs was 22, 9 and 7 for Termination, Indemnity and Insurance respectively. The hyper-parameter choice for paragraph vectors and classifier models are cross validated by comparing end to end performance of predicting the categories. Vector representation built using Distributed memory architecture; trained using negative sampling (Table I) with number of samples to be updated (K)=10 (Table II), subsampling threshold=$10^{-6}$ (Table III), context window=10 (Table IV), vector size=100 (Table V); and classified using linear SVM (Table VII) outperforms all other combinations. All words are converted to lower case. We have a vocabulary of 1,442 words. Special characters are treated as normal words.

The performance of individual risk category classifier model is further tuned by updating the C-values. C value of 1 works best for both Indemnity and Termination (Table VIII). Thus by parameter tuning, the performance of the overall model has increased by 5% points (Table I, Table VIII) in terms of accuracy, for the risk category "Termination". The higher values of precision, recall and F1-score for "Termination" and "Indemnity" as compared to "Insurance" (Table VIII) are driven by a larger training data. The poor performance of the risk category, "Insurance" (trained on <100 paragraphs) suggests training on a larger dataset to produce reliable results. Also contextualization of "Insurance" with associated words to assess risk was not clear, as we found out by manual review of dictionary and training set. Thus performance of this framework is largely dependent on the size of the training data and propriety of pre-defined risk category and association. Using these selections, the test results for Indemnity and Termination are displayed in Table IX.

The varied functionalities of our framework is detailed in Table X with a sample screen-shot in Figure 2.

TABLE X. FRAMEWORK FUNCTIONALITY DESCRIPTION

| Framework Functionalities | Description |
|---|---|
| 1 | Document Upload |
| 2 | Document Repository |
| 3 | Selection of Risk Tokens |
| 4 | Extracted risk prone paragraphs for the risk category selected |
| 5 | Probability Values |
| 6 | Reviewing and Commenting |
| 7 | Feedback to decline an identified paragraph |
| 8 | Original legal document |
| 9 | Export reports |

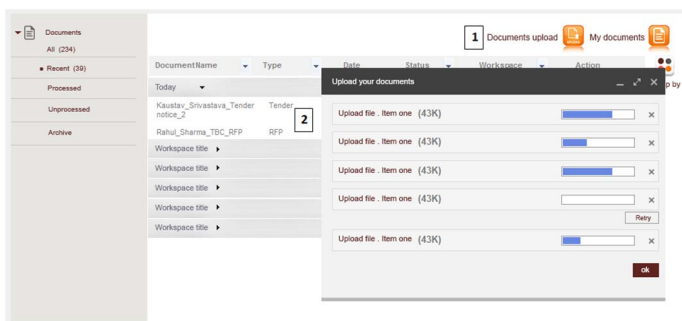

Fig.2."Risk-o-Meter" framework: Document Upload and Document Repository

## VI. FUTURE WORK

The current framework considers all risk prone paragraphs and their associated categories with equal importance. E.g. Statement 1 for Indemnity: "The Consultant shall at all times indemnify and keep indemnified the Company against all claims/damages etc. for any infringement of any Intellectual Property Rights (IPR) while providing its services under the Project." Statement 2 for Termination: "If the firm stands dissolved /reconstituted and the name/ style of the firm is changed." The current framework would identify both the paragraphs as risky but would not differentiate between their severities. To further aid legal professionals in the task of analyzing risk reward ratio, we propose to incorporate a module which considers the severity and impact of different risk paragraphs; and superimpose with risk emanating from associated categories. We will also ascertain the distance of association to weight the risk quotient as necessary. Thus they could be classified as High, Medium or Low; or their impact rated on a scale of 1-5, with 5 being the highest. Based on the severity and impact of the identified risk category, the framework could also be trained for suggesting probable mitigation options.

## VII. CONCLUSION

The time and cost involved in manual assessment of legal documents clearly indicate the need for developing an AI-based system that makes risk analysis of contracts fast, error free and person independent; so that the decision to accept/reject/mitigate to tolerable limit is made easy. In this paper, we presented our framework, "risk-o-meter" which has reduced the average time taken by our legal professionals in reviewing and assessing the risk in legal documents. It thus fosters a risk-aware environment for sustainable growth and knowledgeable decision making for the organization.

The framework can be tailored for uncovering relevant information from multiple kinds of documents by optimizing the hyper-parameters based on the quality and quantity of the available data. Another potential application of the framework could be for processing claims documents wherein accident descriptions could be used for better understanding of the accident scenario and accordingly predicting the accident types such as bodily damage, property damage and others. Our framework, thus provides a scalable, efficient and reliable solution for reviewing and assessing all kinds of documents.